\documentclass[accepted]{uai2021} % after acceptance, for a revised
\usepackage[american]{babel}
\usepackage{natbib} % has a nice set of citation styles and commands
\usepackage{mathtools} % amsmath with fixes and additions
\usepackage{booktabs} % commands to create good-looking tables
\usepackage{tikz} % nice language for creating drawings and diagrams

\usepackage{url}            % simple URL typesetting
\usepackage{amsfonts}       % blackboard math symbols
\usepackage{nicefrac}       % compact symbols for 1/2, etc.
\usepackage{microtype}      % microtypography
\usepackage{makecell}
\usepackage{graphicx}
\usepackage{adjustbox}
\usepackage{sidecap}
\usepackage{subcaption}
\usepackage{amsmath}
\usepackage{wrapfig}
\usepackage{lineno}
\usepackage{comment}
\usepackage{amssymb}
\usepackage{times}
\usepackage{epsfig}
\usepackage{enumitem}
\title{The Unreasonable Effectiveness of the \\ Final Batch Normalization Layer}

\author[1]{\href{mailto:Veysel Kocaman <v.kocaman@liacs.leidenuniv.nl>?Subject=Your UAI 2021 paper}{Veysel Kocaman}{}} % Lead author
\author[2]{Ofer M. Shir}
\author[1]{Thomas Bäck}
\affil[1]{%
    LIACS\\
    Leiden University\\
    Leiden, The Netherlands
}
\affil[2]{%
    Computer Science Department\\
    Tel-Hai College and Migal Institute\\
    Upper Galilee, Israel
}

\begin{document}
\maketitle

\begin{abstract}

%\todo{vk -another title suggestion: The Unreasonable Effectiveness of the Final Batch Normalization Layer (refering famous works having similar titles with 'Unreasonable')}

Early-stage disease indications are rarely recorded in real-world domains, such as Agriculture and Healthcare, and yet, their accurate identification is critical in that point of time.
In this type of highly imbalanced classification problems, which encompass complex features, deep learning (DL) is much needed because of its strong detection capabilities. 
At the same time, DL is observed in practice to favor majority over minority classes and consequently suffer from inaccurate detection of the targeted early-stage indications. 
In this work, we extend the study done by ~\citep{kocaman2020improving}, showing that the final BN layer, when placed before the softmax output layer, has a considerable impact in highly imbalanced image classification problems as well as undermines the role of the softmax outputs as an uncertainty measure. This current study addresses additional hypotheses and reports on the following findings: 
(i) the performance gain after adding the final BN layer in highly imbalanced settings could still be achieved after removing this additional BN layer in inference; 
(ii) there is a certain threshold for the \textit{imbalance ratio}
upon which the progress gained by the final BN layer reaches its peak; 
(iii) the batch size also plays a role and affects the outcome of the final BN application; 
(iv) the impact of the BN application is also reproducible on other datasets and when utilizing much simpler neural architectures;
(v) the reported BN effect occurs only per a single majority class and multiple minority classes -- i.e., no improvements are evident when there are two majority classes; and finally, (vi) utilizing this BN layer with sigmoid activation has almost no impact when dealing with a strongly imbalanced image classification tasks.
\end{abstract}

\section{Introduction}
\label{sec:introduction}

Detecting anomalies that are hardly distinguishable from the majority of observations is a challenging task that often requires strong learning capabilities since anomalies appear scarcely, and in instances of diverse nature, a labeled dataset representative of all forms is typically unattainable. 
%Despite tremendous advances in computer vision and object recognition algorithms, their effectiveness remains strongly dependent upon the size and distribution of the training set. 
Despite tremendous advances in computer vision and object recognition algorithms in the past few years, their effectiveness remains strongly dependent upon the datasets' size and distribution, which are usually limited under real-world settings.
This work is mostly concerned with hard classification problems at early-stages of abnormalities in certain domains (i.e. crop, human diseases, chip manufacturing), which suffer from lack of data instances, and whose effective treatment would make a dramatic impact in these domains. For instance, fungus's visual cues on crops in agriculture or early-stage malignant tumors in the medical domain are hardly detectable in the relevant time-window, while the highly infectious nature leads rapidly to devastation in a large scale. Other examples include detecting the faults in chip manufacturing industry, automated insulation defect detection with thermography data, assessments of installed solar capacity based on earth observation data, and nature reserve monitoring with remote sensing and deep learning. However, class imbalance poses an obstacle when addressing each of these applications.

In recent years, reliable models capable of learning from small samples have been obtained through various approaches, such as autoencoders~\citep{beggel2019robust}, class-balanced loss (CBL) to find the effective number of samples required~\citep{cui2019class}, fine tuning with transfer learning~\citep{hussain2018study}, data augmentation~\citep{shorten2019survey}, cosine loss utilizing (replacing categorical cross entropy)~\citep{barz2020deep}, or prior knowledge~\citep{lake2015human}. 

\begin{figure*}[!hbt]
\begin{center}
\includegraphics[width=0.9\textwidth]{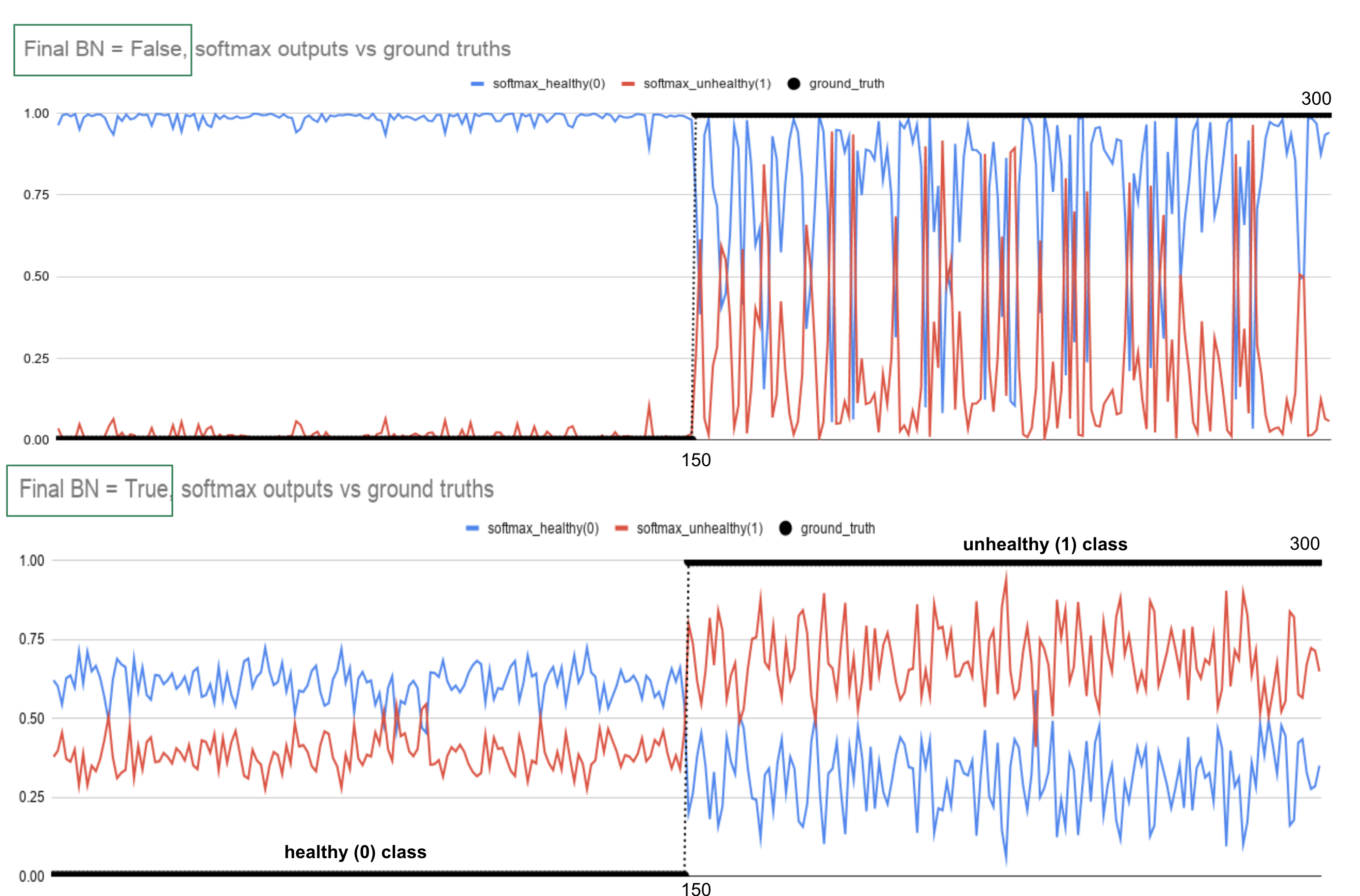}
\end{center}
\caption{The $x$-axis represents the ground truth for all 150 healthy (0) and 150 unhealthy (1) images in the test set while red and blue lines represent final softmax output values between 0 and 1 for each image. \textbf{Top chart (without final BN):} When ground truth (black) is $\text{class} = 0$ (healthy), the softmax output for $\text{class} = 0$ is around 1.0 (blue, predicting correctly). But when ground truth (black) is $\text{class} = 1$ (unhealthy), the softmax output for $\text{class} = 1$ (red points) changes between 0.0 and 1.0 (mostly below 0.5, NOT predicting correctly). \textbf{Bottom chart (with final BN):} When ground truth (black) is $\text{class} = 0$ (healthy), the softmax output is between 0.5 and 0.75 (blue, predicting correctly). When ground truth (black) is $\text{class} = 1$ (unhealthy), the softmax output  (red points) changes between 0.5 and 1.0 (mostly above 0.5, predicting correctly).}
\label{fig:softmax_outputs}
\end{figure*}

~\citep{kocaman2020improving} presented an effective modification in the neural network architecture, with a surprising simplicity and without a computational overhead, which enables a substantial improvement using a smaller number of anomaly samples in the training set. 
The authors empirically showed that the final BN layer before the softmax output layer has a considerable impact in highly imbalanced image classification problems. They reported that under artificially-generated skewness of 99\% vs.~1\% in the PlantVillage (PV) image dataset~\citep{mohanty2016using}, the initial F1 test score increased from the 0.29-0.56 range to the 0.95-0.98 range (almost triple) for the minority class when BN modification applied. They also argued that, a model might perform better even if it is not confident enough while making a prediction, hence the softmax output may not serve as a good uncertainty measure for DNNs (see Figure~\ref{fig:softmax_outputs}). 

This shows that DNNs have the tendency of becoming ‘over-confident’ in their predictions during training, and this can reduce their ability to generalize and thus perform as well on unseen data. In addition, large datasets can often comprise incorrectly labeled data, meaning inherently the DNN should be a bit skeptical of the ‘correct answer’ to avoid being overconfident on bad answers. This was the main motivation of M{\"u}ller et al.~\citep{muller2019does} for proposing the \textit{label smoothing}, a loss function modification that has been shown to be effective for training DNNs. Label smoothing encourages the activations of the penultimate layer to be close to the template of the correct class and equally distant to the templates of the incorrect classes~\citep{muller2019does}. Despite its relevancy, ~\citep{kocaman2020improving} also reports that label smoothing did not do well in their study as previously mentioned by Kornblith et al.~\citep{chelombiev2019adaptive} who demonstrated that label smoothing impairs the accuracy of transfer learning, which similarly depends on the presence of non-class-relevant information in the final layers of the network.

In this study, we extend previous efforts done by ~\citep{kocaman2020improving} to devise an effective approach to enable learning of minority classes, given the surprising evidence of applying the final Batch Normalization (BN) layer. 
%Our primary research question was the following: \textit{What is the most effective approach to enable learning of minority classes, could Batch Normalization (BN) serve as one?}

Given these recent findings, we formulate and test additional hypotheses and report our observations in what follows. The concrete contributions of this paper are the following:

\begin{itemize}

    \item  The performance gain after adding the final BN layer in highly imbalanced settings could still be achieved after removing this additional BN layer during inference; in turn enabling us to get a performance boost with no additional cost in production.

    \item There is a certain threshold for the ratio of the imbalance for this specific PV dataset, upon which the progress is the most obvious after adding the final BN layer.

    \item The batch size also plays a role and significantly affects the outcome.

    \item  We replicated the similar imbalanced scenarios in MNIST dataset, reproduced the same BN impact, and furthermore demonstrated that the final BN layer has a considerable impact not just in modern CNN architectures but also in simple CNNs and even in one-layered feed-forward fully connected (FC) networks.

    \item We illustrate that the performance gain occurs only when there is a single majority class and multiple minority classes; and no improvement observed regardless of the final BN layer when there are two majority classes. 

    \item We argue that using the final BN layer with sigmoid activation has almost no impact when dealing with a strongly imbalanced image classification tasks.
\end{itemize}

The remainder of the paper is organized as follows: 
Section~\ref{sec:background} gives some background concerning the role of the BN layer in neural networks.
Section~\ref{sec:summary_previous} summarizes the previous findings and existing hypotheses in the previous work done by ~\citep{kocaman2020improving} and then lists the derived hypotheses that will be addressed throughout this study.
Section~\ref{sec:ImplementationDetails} elaborates the implementation details and settings for our new experiments and presents results.
Section~\ref{sec:discussion}  discusses the findings and proposes possible \textit{mechanistic} explanations.
Section~\ref{sec:conclusion} concludes this paper by pointing out key points and future directions.

\section{Background}
\label{sec:background}
In order to better understand the novel contributions of this study, 
in this section, we give some background information about the Batch Normalization (BN) ~\citep{ioffe2015batch} concept. Since the various applications of BN in similar studies and related work have already been investigated thoroughly in our previous work~\citep{kocaman2020improving}, we will only focus on the fundamentals of BN in this chapter. 

Training deep neural networks with dozens of layers is challenging as the networks can be sensitive to the initial random weights and configuration of the learning algorithm. One possible reason for this difficulty is that the distribution of the inputs to layers deep in the network may change after each mini-batch when the weights are updated. This slows down the training by requiring lower learning rates and careful parameter initialization, makes it notoriously hard to train models with saturating nonlinearities~\citep{ioffe2015batch}, and can cause the learning algorithm to forever chase a moving target. This change in the distribution of inputs to layers in the network is referred to by the technical name \textit{“internal covariate shift”} (ICS).

BN is a widely adopted technique that is designed to combat ICS and to enable faster and more stable training of deep neural networks (DNNs). It is an operation added to the model before activation which normalizes the inputs and then applies learnable scale ($\gamma$) and shift ($\beta$) parameters to preserve model performance. Given $m$ activation values $x_1\ldots,x_m$ from a mini-batch ${\cal B}$ for any particular layer input $x^{(j)}$ and any dimension $j \in \{1,\ldots,d\}$, the transformation uses the mini-batch mean 
$\mu_{\cal B}= 1/m \sum_{i=1}^m x_i$ and variance $\sigma_{\cal B}^2 = 1/m \sum_{i=1}^m (x_i - \mu_{\cal B})^2$ for normalizing the $x_i$ according to $\hat{x}_i = (x_i - \mu_{\cal B})/\sqrt{\sigma_{\cal B}^2 + \epsilon}$
%
\begin{comment}
\begin{linenomath*}
\[\hat{x}_i = \frac{x_i - \mu_{\cal B}}{\sqrt{\sigma_{\cal B}^2 + \epsilon}}\nonumber\]
\end{linenomath*}
\end{comment}
%
and then applies the scale and shift to obtain the transformed values $y_i$ = $\gamma \hat{x}_i + \beta$. The constant $\epsilon > 0$ assures numerical stability of the transformation.

%\tb{Isn't it setting the mean and standard deviation to values re-centered and re-scaled according the empirical mean and standard deviation of the batch? That is different from what you say above.}\tb {(vk) it uses the std dev and mean from the batch but also uses two learnable parameter to shift (add a bias)}

BN has the effect of stabilizing the learning process and dramatically reducing the number of training epochs required to train deep networks; and using BN 
%\tb{Can we agree on one notation, either Batch Norm or normalization? I understand these are identical, you use Batch Norm as a shortcut. And we also have Batch-Norm and Batch Normalization ...}\tb {(vk) great point! I will fix in the entire draft}
makes the network more stable during training. This may require the use of much larger learning rates, which in turn may further speed up the learning process. 

Though BN has been around for a few years and has become common in deep architectures, it remains one of the DL concepts that is not fully understood, having many studies discussing why and how it works.
Most notably, Santurkar et al.~\citep{santurkar2018does} recently demonstrated that such distributional stability of layer inputs has little to do with the success of BN and the relationship between ICS and BN is tenuous. Instead, they uncovered a more fundamental impact of BN on the training process: it makes the optimization landscape significantly smoother. This smoothness induces a more predictive and stable behavior of the gradients, allowing for faster training. Bjorck et al.~\citep{bjorck2018understanding} also makes similar statements that the success of BN can be explained without ICS. They argue that being able to use larger learning rate increases the implicit regularization of the gradient, which improves generalization. 
 
Even though BN adds an overhead to each iteration (estimated as additional 30\% computation~\citep{mishkin2015all}), the following advantages of BN outweigh the overhead shortcoming:

    \begin{itemize}
    \item It improves gradient flow and allows training deeper models (e.g., ResNet).
    \item It enables using higher learning rates because it eliminates outliers' activation, hence the learning process may be accelerated using those high rates.
    \item It reduces the dependency on initialization and then reduces overfitting due to its minor regularization effect. 
Similarly to dropout, it adds some noise to each hidden layer’s activation.
    \item Since the scale of input features would not differ significantly, the gradient descent may reduce the oscillations when approaching the optimum and thus converge faster.
    \item BN reduces the impacts of earlier layers on the following layers in DNNs. 
Therefore, it takes more time to train the model to converge. However, the use of BN can reduce the impact of earlier layers by keeping the mean and variance fixed, which in some way makes the layers independent from each other. 
Consequently, the convergence becomes faster.
\end{itemize}

%The development of BN as a normalization technique was a turning point in the development of DL models, and it enabled various networks to train and converge. Despite its great success, BN exhibits drawbacks that are caused by its distinct behavior of normalizing along the batch dimension. One of the major disadvantages of BN is that it requires sufficiently large batch sizes to obtain good results.  This prevents the user from exploring higher-capacity models that would be limited by memory. To solve this problem, several other normalization variants are developed, such as Layer Normalization (LN)~\cite{ba2016layer}, Instance Normalization (IN)~\cite{ulyanov2016instance}, Group Normalization (GN)~\cite{wu2018group} and Filter Response Normalization (FRN)~\cite{singh2019filter}.
\section{Previous Findings and Existing Hypotheses } \label{sec:summary_previous}
%Detecting anomalies that are hardly distinguishable from the majority of observations is a challenging task that often requires strong learning capabilities.  
%While researching about the most effective approach to enable learning of minority classes, 
In the work done by \citep{kocaman2020improving}, the authors focused their efforts on the role of BN layer in DNNs, where in the first part of the experiments ResNet34~\citep{simonyan2014very} and VGG19 CNN architectures~\citep{he2016deep} are utilized. They first addressed the complete PV original dataset and trained a ResNet34 model for 38 classes. Using scheduled learning rates~\citep{smith2017cyclical}, they obtained 99.782\% accuracy after 10 epochs -- slightly improving the PV project's record of 99.34\% when employing  GoogleNet~\citep{mohanty2016using}. 
In what follows, we summarize the previous observations borrowed from \citep{kocaman2020improving} and then formulate the derived hypotheses that became the core of the current study.

\subsection{Adding a Final Batch Norm Layer Before the Output Layer}
By using the imbalanced datasets for certain plant types (1,000/10 in the training set, 150/7 in the validation set and 150/150 in the test set), \citep{kocaman2020improving} performed several experiments with the VGG19 and ResNet34 architectures.
The selected plant types were Apple, Pepper and Tomato - being the only datasets of sufficient size to enable the 99\%-1\% skewness generation. All the tests are run with batch size 64.

In order to fine-tune the network for the PV dataset, the final classification layer of CNN architectures is replaced by Adaptive Average Pooling (AAP), BN, Dropout, Dense, ReLU, BN and Dropout followed by the Dense and BN layer again. The last layer of an image classification network is often a FC layer with a hidden size being equal to the number of labels to output the predicted confidence scores that are normalized by the softmax operator to obtain predicted probabilities. In their implementation, they added another 2-input BN layer after the last dense layer (before softmax output) in addition to existing BN layers in the tail and 4 BN layers in the head of the DL architecture (e.g., ResNet34 possesses a BN layer after each convolutional layer, having altogether 38 BN layers given the additional 4 in the head).

At first they run experiments with VGG19 architectures for selected plant types by adding the final BN layer. When they train this model for 10 epochs and repeat this for 10 times, they observed that the F1 test score is increased from 0.2942 to 0.9562 for unhealthy Apple, from 0.7237 to 0.9575 for unhealthy Pepper and from 0.5688 to 0.9786 for unhealthy Tomato leaves. They also achieved significant improvements in healthy samples (being the majority in the training set). See Table~\ref{tab:three_plants} for details.

\begin{table*}[bt!]
\caption{Averaged \textbf{F1 test set} performance values over 10 runs, alongside BN's total improvement, using 10 epochs with VGG19, with/without BN and with Weighted Loss (WL) without BN.}
\label{tab:three_plants}
\centering
\begin{tabular}{llcccc}
\hline
plant & class & \multicolumn{1}{c}{\thead{without final\\ BN}} & \thead{with WL\\ (no BN)} & \multicolumn{1}{c}{\thead{with final BN\\ (no WL)}} & \thead{BN total\\ improvement} \\
\hline
Apple  & Unhealthy & \textbf{0.2942} & \textbf{0.7947} & \textbf{0.9562} & \textbf{0.1615}\\
       & Healthy   & 0.7075 & 0.8596 & 0.9577 & 0.0981\\ \hline
Pepper & Unhealthy & \textbf{0.7237} & \textbf{0.8939} & \textbf{0.9575} & \textbf{0.0636} \\ 
       & Healthy   & 0.8229 & 0.9121 & 0.9558 & 0.0437\\ \hline
Tomato & Unhealthy & \textbf{0.5688} & \textbf{0.8671} & \textbf{0.9786} & \textbf{0.1115}\\
       & Healthy   & 0.7708 & 0.9121 & 0.9780 & 0.0659\\
\hline
\end{tabular}
\end{table*}

\subsection{Experimentation on PlantVillage Dataset subject to Different Configurations}
Using the following six configuration variations with two options each, the authors created 64 different configurations which they tested with ResNet34 (training for 10 epochs only):
%and then trained the same model for 10 epochs with ResNet34. Since VGG19 is computationally more expensive than ResNet34, we run the next experiments only in ResNet34: 
Adding (\checkmark) a final BN layer just before the output layer (BN), using (\checkmark) weighted cross-entropy loss~\citep{goodfellow2016deep} according to class imbalance (WL), using (\checkmark) data augmentation (DA), using (\checkmark) mixup (MX)~\citep{zhang2017mixup}, unfreezing  (\checkmark) or freezing (learnable vs pre-trained weights) the previous BN layers in ResNet34 (UF), and using (\checkmark) weight decay (WD)~\citep{krogh1992simple}. 
Checkmarks (\checkmark) and two-letter abbreviations are used in Table~\ref{best_scores} to denote configurations. When an option is disabled across all configurations, its associated column is dropped.
%
%\vspace{-0.5cm}
\begin{comment}
\begin{table*}[tbp!]
\caption{Best performance metrics over the Apple dataset under various configurations using ResNet34.}
\label{best_scores}
%\scalebox{0.77}{
\centering
\begin{tabular}{ccccccccccccc}\hline
%{l *{8}{p{1cm}}
\multicolumn{1}{l}{Class} &
  \thead{Config\\ Id} &
  \thead{Test set\\ precision} &
  \thead{Test set\\  recall} &
  \thead{Test set\\ F1-score} &
  Epoch &
%  \thead{With BN\\ final} &
    BN &
%  \thead{Weighted\\ loss} &
    WL &
% \thead{Data\\ augment} &
    DA &
% Mixup &
    MX &
%  \thead{Train BN\\ Layers} &
    UF &
% \thead{Weight\\ Decay} \\ 
    WD \\
\hline
Unhealthy & \textbf{31} & 0.9856 & 0.9133 & \textbf{0.9481} & 6 & \checkmark  & & & & & \\ 
          & 23 & 0.9718 & 0.9200 & 0.9452 & 6 & \checkmark  & & \checkmark  & & & \\ 
          & 20 & 0.9926 & 0.8933 & 0.9404 & 7 & \checkmark  & & \checkmark  & & \checkmark  & \checkmark \\ \hline
Healthy   & 31 & 0.9193 & 0.9867 & 0.9518 & 6 & \checkmark  & & & & & \\ 
          & 23 & 0.9241 & 0.9733 & 0.9481 & 6 & \checkmark  & & \checkmark  & & & \\ 
          & 20 & 0.9030 & 0.9933 & 0.9460 & 7 & \checkmark  & & \checkmark  & & \checkmark  & \checkmark \\ 
\hline
\end{tabular}
%}
\end{table*}
\end{comment}

\begin{table*}[bt!]
\caption{Best performance metrics over the Apple dataset under various configurations using ResNet34.}
\label{best_scores}
%\scalebox{0.77}{
\centering
\begin{tabular}{lccccccccccc}
%\addtolength{\tabcolsep}{3pt}
\hline
%{l *{8}{p{1cm}}
{Class} &
  \thead{Config\\ Id} &
  \thead{Test set\\ precision} &
  \thead{Test set\\  recall} &
  \thead{Test set\\ F1-score} &
  Epoch &
%  \thead{With BN\\ final} &
    BN &
%  \thead{Weighted\\ loss} &
%    WL &
% \thead{Data\\ augment} &
    DA &
% Mixup &
%    MX &
%  \thead{Train BN\\ Layers} &
    UF &
% \thead{Weight\\ Decay} \\ 
    WD \\
\hline
Unhealthy & \textbf{31} & 0.9856 & 0.9133 & \textbf{0.9481} & 6 & \checkmark  & & &  \\ 
($\text{class}=1$) & 23 & 0.9718 & 0.9200 & 0.9452 & 6 & \checkmark  & \checkmark  & & \\ 
          & 20 & 0.9926 & 0.8933 & 0.9404 & 7 & \checkmark  & \checkmark  & \checkmark  & \checkmark \\ \hline
\end{tabular}
%}
\end{table*}

As shown in Table~\ref{best_scores}, just adding the final BN layer was enough to get the highest F1 test score in both classes. Surprisingly, although there is already a BN layer after each convolutional layer in the backbone CNN architecture, adding one more BN layer just before the output layer boosts the test scores. 
%\tb{That is not there anymore, right?}
%\tb {(vk) I talk about higher losses here but I had to drop loss columns from the table above due to cosmetic purposes. We may need to add again} 
%when we have the highest validation and test accuracies (lowest test error rate). That is, our model is performing well but less confident due to high losses. This is mainly because we train the models with a smaller number of epochs and do not let the model converge further.
%Actually, training the models faster is one of the expected behaviors of using BN but it should have caused lower losses as well. 
Notably, the 3rd best score (average score for configuration 31 in Table~\ref{best_scores}) %\tb{3rd best score? All I want to say here is that you maybe mean the "3rd best score"? "best 3rd does not make sense". Also, say again what you refer to, where I can find this number, etc.} 
%\tb {(vk) I just pointe the number.. when we turn of all the other params, and add final BN, we have 0.9588; when we add unfreeze the previous BN layers and make them trainable, we get 0.9633. so it adds a little bit boost but all the configs in top 3 have final BN true} 
%\tb{I understand things better now but you have to explain what your table column headings mean and how they relate to the text! For example, train-bn and true-wd and so on. It is not even clear that average refers to the average value of aspple, pepper, tomato, I assume, and which performance measure the table shows - F1 I assume.}\tb {(vk) I added the column names next to explanations in the paragraph above the table ?}
is achieved just by adding a single BN layer before the output layer, even without unfreezing the previous BN layers. 

%
%It is also worth mentioning that the effect of weighted loss is considerably high. Just by using weighted loss (0.01 vs 1.0) in the cross entropy function, we nearly doubled the test F1 score for minority class (unhealthy) (from 0.42 to 0.81). (I dropped this statement as we have no space to explain this and it is also out of scope)
%
%
%
%

One of the important observations is that the model without the final BN layer is pretty confident even if it predicts falsely. 
But the proposed model with the final BN layer predicts correctly even though it is less confident. They basically ended up with less confident but more accurate models in less than 10 epochs. %The classification probabilities for unhealthy (class-1) class from both versions (bn\_final = False vs True) are shown in Table~\ref{tab:confident_wrongs}.
The classification probabilities for five sample images from the unhealthy class ($\text{class}=1$) with final BN layer (right column) and without final BN layer (left column) are shown in Table~\ref{tab:confident_wrongs}.
As explained above, without the final BN layer, these anomalies are all falsely classified (recall that ${\cal P}_{\text{softmax}}(\text{class} = 0) = 1 - {\cal P}_{\text{softmax}}(\text{class} = 1)$).

\begin{table}[htbp]
\caption{Softmax output values (representing class probabilities) for five sample images of unhealthy %($\text{class}=1$) 
plants. Left column: Without final BN layer, softmax output values for unhealthy, resulting in a wrong classification in each case. Right column: With final BN layer, softmax output value for unhealthy, resulting in correct but less "confident" classifications.}
\label{tab:confident_wrongs}
%\begin{center}
\centering
\begin{tabular}{cc}
\toprule
\multicolumn{1}{c}{Without final BN layer} &
\multicolumn{1}{c}{With final BN layer} \\
%\multicolumn{1}{c}{${\cal P}_{\text{softmax}}(\text{class} = 1)$} &
%\multicolumn{1}{c}{${\cal P}_{\text{softmax}}(\text{class} = 1)$} \\
\midrule
0.1082 & 0.5108\\
0.1464 & 0.6369\\
0.1999 & 0.6082\\
0.2725 & 0.6866\\
0.3338 & 0.7032\\
% 0.1464 & 0.6369  \\
% 0.1999 & 0.6082  \\
% 0.2725 & 0.6866  \\
% 0.3338 & 0.7032  \\ 
\bottomrule
\end{tabular}
%\end{center}
\end{table}
%\end{SCtable}

\subsection{Derived Hypotheses}
\label{ssec:hypotheses}
%\textbf{***TODO: Specify the hypotheses of the current work -- that is, (i)-(vi) of the abstract.}

Under all these observations and findings mentioned above, we derived the following hypotheses for our new study: 

\begin{itemize}
\item The added complexity to the network by adding the final BN layer could be eliminated by removing the final BN layer in inference without compromising the performance gain achieved.
\item There might be a certain level of skewness upon which the progress reaches its peak without further sizing the minority class.
\item Since the trainable parameters in a BN layer also depend on the batch size (i.e., number of samples) in each iteration (mini-batch), its sizing could also play a role on the level of progress with the final BN layer.
\item The observations and performance gain with respect to the PV dataset, upon utilizing ResNet and VGG architectures, may not be reproduced with any other dataset or with much simpler neural architectures.
\item Since the number of units in the output layer depends on the number of classes in the dataset, the performance gain may not be achieved in multi-classification settings and the number of majority and minority classes can affect the role of the final BN layer.
\item Since sigmoid activation is also one of the most widely used activation functions in the output layer for the binary classification problems, the performance gain could be achieved with sigmoid outputs as well.
\end{itemize}

Next, we report on addressing these hypotheses, one by one, and describe our empirical findings in detail.
\section{Implementation Details and Experimental Results}
\label{sec:ImplementationDetails}

\subsection{Removing the additional BN layer during inference (H-1)}

Since adding the final BN layer adds a small overhead (four new parameters) to the network at each iteration, we experimented if the final BN layer could be dropped once the training is finished so that we can avoid the cost. Dropping this final BN layer means that training the network from end to end, and then chopping off the final BN layer from the network before saving the weights. We tested this hypothesis for Apple, Pepper and Tomato images from PV dataset under three conditions with 1\% imbalance ratio: \textit{Without final BN, with final BN and then removing the final BN during testing}. We observed that removing the final BN layer in inference would still give us a considerable boost on minority class without losing any performance gain on the majority class. The results in Table~\ref{tab:removed_bn} show that the performance gain is very close to the configuration in which we used the final BN layer both in training and inference time. 
As a consequence, we confirm hypothesis (H-1) %that we derived in section \ref{ssec:hypotheses} 
by showing that the final BN layer can indeed be removed in inference without compromising the performance gain.

\begin{table*}[htb]
\caption{Training with the final BN layer, and then dropping this layer while evaluating on the test set proved to be still useful in terms of improving the classification score on minority classes, albeit not as much as with the final BN layer kept (imbalance ratio 0.01, epoch 10, batch size 64).}
\label{tab:removed_bn}
\centering
\begin{tabular}{lcccccc}
\hline
 & \multicolumn{2}{c}{Apple} & \multicolumn{2}{c}{Pepper} & \multicolumn{2}{c}{Tomoto} \\\cline{2-7}
 & healthy & unhealthy & healthy & unhealthy &  healthy & unhealthy \\\hline
with no final BN & 0.71 & 0.22 & 0.74 & 0.45 & 0.74 & 0.46 \\
with final BN & 0.92 & 0.91 & 0.94 & 0.94 & 0.98 & 0.98 \\
train with final BN & 0.74 & \textbf{0.83} & 0.75 & \textbf{0.82} & 0.78 & \textbf{0.85}\\
remove while testing & &  &  &  &  & \\\hline
\end{tabular}
\end{table*}

\subsection{Impact level regarding the imbalance ratio (H-2) and the batch size (H-3)}

\begin{figure*}[htbp!]
\begin{subfigure}{0.498\textwidth}
\includegraphics[width=1.0\linewidth]{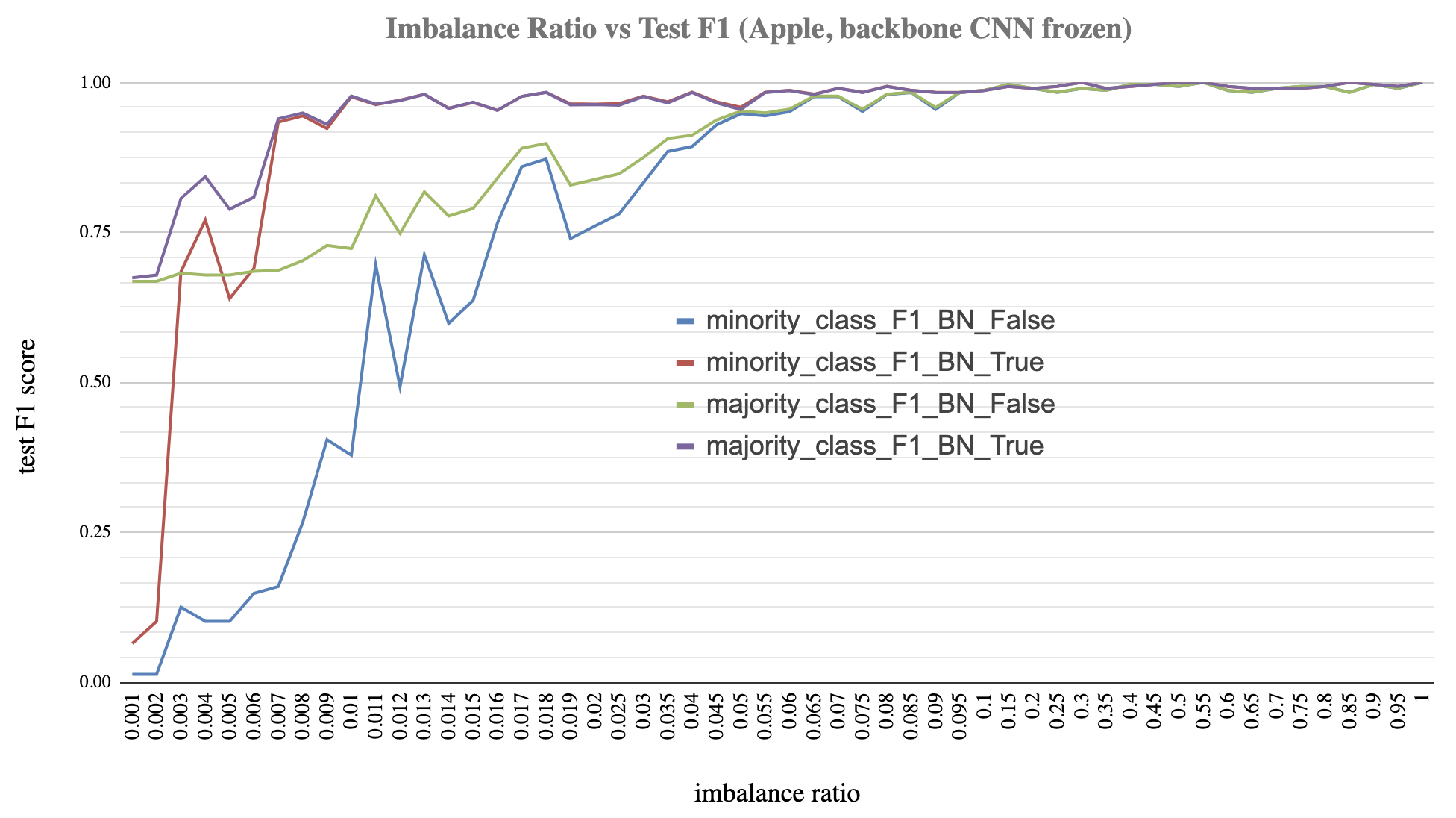}
\subcaptionbox{The impact of final BN layer on the F1 test score of each class for Apple plant. The impact is the most obvious when the ratio of minority class to the majority is less than 0.1.
\label{fig:imbalance_f1_ratio_in}}[0.95\linewidth]

\end{subfigure}
\begin{subfigure}{0.505\textwidth}
\includegraphics[width=1.0\linewidth]{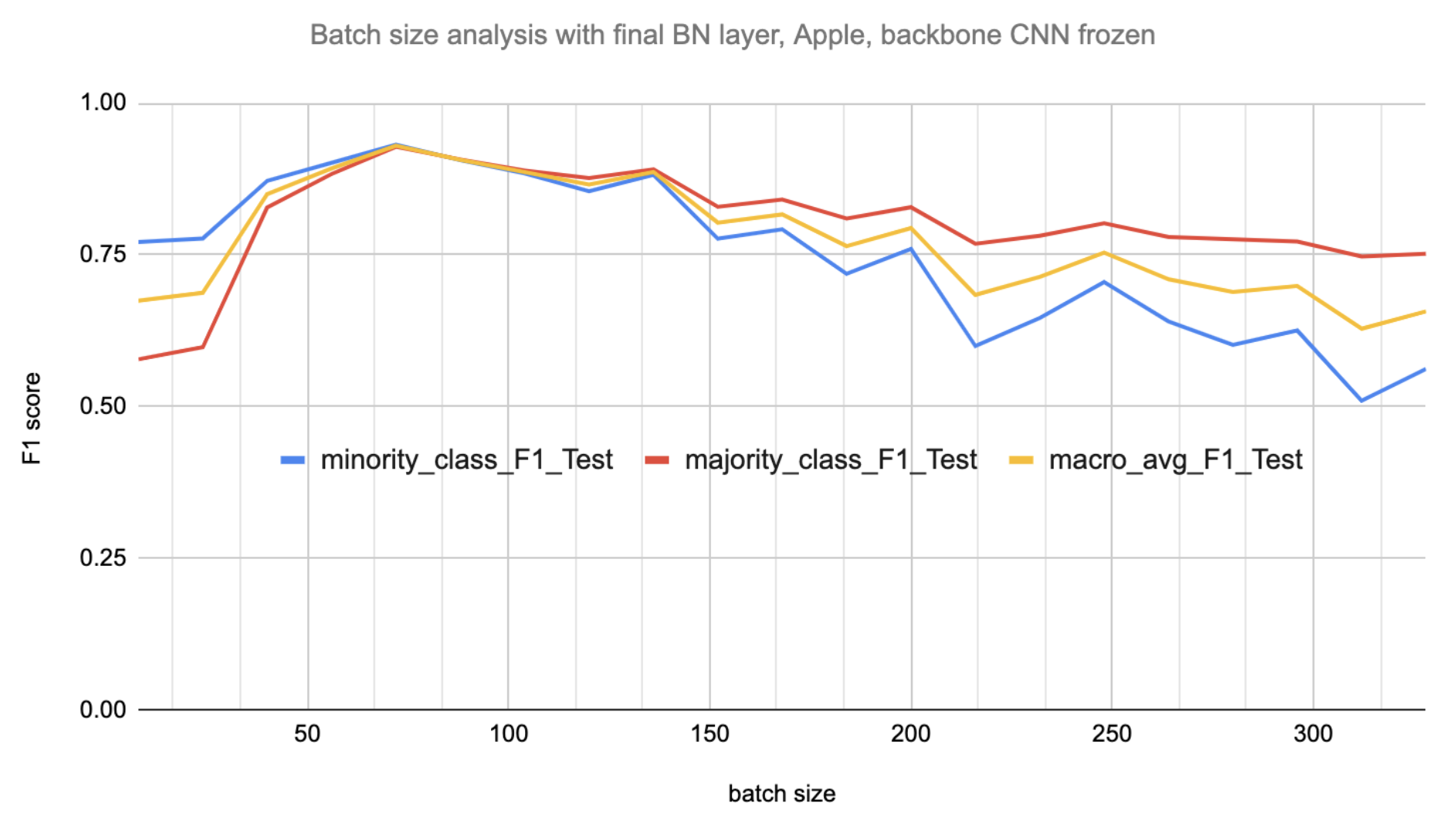}
\subcaptionbox{The impact of batch size on the F1 test score when the final BN layer added.  The highest score is gained when the batch size is 64 and then the accuracy starts declining.\label{fig:bs_analysis}}[0.95\linewidth]

\end{subfigure}
\caption{Imbalance ratio and batch size analysis with respect to the final BN layer added}
\label{fig:imbalance_f1_ratio}
\end{figure*}

~\citep{kocaman2020improving} empirically shows that the final BN layer, when placed be-fore the softmax output layer, has a considerable impact in highly imbalanced image classification problems but they fail to explain the impact of using the final BN layer as a function of level of imbalance in the training set. In order to find if there is a certain ratio in which the impact is maximized, we tested this hypothesis (H-1) over various levels of imbalance ratios and conditions explained below. In sum, we ended up with 430 model runs, each with 10 epochs (basically tested with 5, 10, 15, .. 100 unhealthy vs 1000 healthy samples). During the experiments, we observed that the impact of final BN on highly imbalanced settings is the most obvious when the ratio of minority class to the majority is less than 10\%; above that almost no impact. As expected, the impact of the final BN layer is more obvious on minority class than it is on majority class, albeit the level of impact with respect to the imbalance ratio is almost same, and levels off around 10\%.
It is mainly because of the fact that the backbone architecture (ResNet34) is already good enough to converge faster on such data set (Plant Village) and the model does well on both classes after 10\% imbalance (having more than 100 unhealthy with respect to 1000 healthy samples can already be handled regardless of final BN trick). During these experiments, we also tested if unfreezing the previous layers in the backbone CNN architecture (ResNet34) would also matter. We observed that unfreezing the pretrained layers helps without even final BN layer, but unfreezing adds more computation as the gradient loss will be calculated for each one of them. After adding the final BN layer and freezing the previous pretrained layers, we observed similar metrics and learning pattern as we did with unfreezing but not with final BN layer. This is another advantage of using the final BN that allow us to freeze the previous pretrained layers. The results are displayed as charts in Figure~\ref{fig:imbalance_f1_ratio_in}. 

%We basically tested with 5, 10, 15, .. 100 unhealthy vs 1000 healthy samples for two plant types (Apple and Tomato but just Apple is shown here), and the results looks quite similar across the different plants even though there are more variation in the unhealthy Tomato samples (there are 9 different blight types in Tomato while there are 3 in Apple). As it is clear from the chart,

%\subsection{The impact of batch size on the effectiveness of final BN trick}

We also experimented if the batch size would also be an important parameter for the minority class test accuracy when the final BN is added and found out that the highest score is gained when the batch size is around 64, whereas the accuracy drops afterwards with larger batches. 
The results are exhibited in Figure~\ref{fig:bs_analysis}. 
Consequently, given the reported observations, we confirm hypotheses (H-2) and (H-3). %, which we derived in section \ref{ssec:hypotheses}, 
%by showing that there might be a certain level of skewness upon which the progress reaches its peak, and that the batch size could also affect the level of progress with the final BN layer.

\subsection{Experimentation on the MNIST Dataset: The impact of final BN layer in basic CNNs and FC networks [(H-4),(H-5),(H-6)]}
In order to reproduce equivalent results on a well known benchmark dataset when utilizing different DL architectures, we set up two simple DL architectures: a CNN network with five Conv2D layers and a one-layer (128-node) FC feed forward NN. Then we sampled several pairs of digits (2 vs 8, 3 vs 8, 3 vs 5 and 5 vs 8) from MNIST dataset that are mostly confused in a digit recognition task due to similar patterns in the pixels. During the experiments, the ratio of minority class to majority is kept as 0.1 and experiments with the simple CNN architecture indicates that, after adding the final BN layer, we gain $\sim$ 20\% boost in minority class and $\sim$ 10\% in majority class (see Table~\ref{tab:Simple_CNN_MNIST_3_8}). Lower standard deviations across the runs with final BN layer also indicates the regularization effect of using the final BN layer. As a result, we rejected the fourth hypothesis (H-4) that we derived in section \ref{ssec:hypotheses} by showing that the performance gain through the final BN layer can be reproduced with another dataset or with much simpler neural architectures.
Another important finding is that the final BN layer boosts the minority class' F1 scores only when there is a single majority class. When two majority classes are set, no improvements are evident regardless of the usage of the final BN layer (see 4th and 5th settings in Table~\ref{mnist_experiments}). Therefore, we partially confirm hypothesis (H-5) %that we derived in section \ref{ssec:hypotheses} 
by showing that the performance gain through the final BN layer can also be achieved in multi-classification settings but the number of majority and minority classes may affect the role of the final BN layer.
In the experiments with a one-layer FC network, we enriched the scope and also tested whether using different loss functions and output activation functions would have an impact on model performance using final BN layer with or without another BN layer after the hidden layer. 
We observed that adding the final BN layer, softmax output layer and categorical crossentropy (CCE) as a loss function have the highest test F1 scores for both classes. It is also important to note that we observe no improvement even after adding the final BN layer when we use sigmoid activation in the output layer (Table~\ref{mnist_experiments}). 
Accordingly, we reject hypothesis (H-6). %that we derived in section \ref{ssec:hypotheses} by showing that the performance gain through the final BN layer cannot be achieved with sigmoid outputs as well.

\begin{table}[htb]
\caption{Average test metrics with a simple CNN network (5xConv2D) to classify 3 (minority) and 8 (majority) images from MNIST dataset with 0.01 imbalance ratio, 10 runs, and 20 epoch per run. }
\label{tab:Simple_CNN_MNIST_3_8}
\centering
\resizebox{0.95\columnwidth}{!}{
\begin{tabular}{lcccc}
          & \thead{with BN\\ minority} & \thead{without BN\\ minority} & \thead{with BN\\ majority} & \thead{without BN\\ majority} \\\hline
Test F1   & 0.9299                                      & 0.7378                                         & 0.9447                                      & 0.8354                                         \\
Std. Dev. & 0.0791                                      & 0.1126                                         & 0.0487                                      & 0.0506\\\hline     
\end{tabular}
}
\end{table}

\begin{table*}[htbp]
\caption{Using the same ResNet-34 architecture and skewness (1\% vs 99\%) and setting up five different configurations across various confusing classes from MNIST dataset, it is clear that adding the final BN layer boosts the minority class F1 scores by 10\% to 30\% only when there is a single majority class. When we have two majority classes, there is no improvement observed regardless of the final BN layer is used or not. (\checkmark) indicates minority classes.}
\label{mnist_experiments}
\centering
%\resizebox{0.95\textwidth}{!}{
\begin{tabular}{lcccccccccccc}
\hline
 & \multicolumn{2}{c}{Setting-1} & \multicolumn{2}{c}{Setting-2} & \multicolumn{2}{c}{Setting-3} & \multicolumn{3}{c}{Setting-4} & \multicolumn{3}{c}{Setting-5} \\
 & 3 (\checkmark)  & 8 & 2 (\checkmark)  & 8 & 3 (\checkmark)  & 5 & 3 (\checkmark) & 5 & 8 & 3 (\checkmark) & 5 (\checkmark) & 8 \\
 \hline
without final BN & 0.19 & 0.69 & 0.25 & 0.70 & 0.24 & 0.70 & 0.06 & 0.77 & 0.78 & 0.28 & 0.32 & 0.56 \\
with final BN & 0.55 & 0.76 & 0.50 & 0.74 & 0.54 & 0.76 & 0.00 & 0.77 & 0.78 & 0.58 & 0.59 & 0.69\\\hline
\end{tabular}%}
%\vspace{-8mm}
\end{table*}

\begin{table}[htbp]
\caption{Using one-layer (128 node) NN and the 0.01 skewness ratio, with nine different settings for 3 (minority) and 8 (majority) classes from MNIST dataset (100-epoch). Adding the final BN layer, softmax output layer and CCE as a loss function has the highest test F1 scores for both classes. (CCE - categorical cross entropy, BCE - binary cross entropy, first BN - a BN layer after the hidden layer).}
\label{imb_mnist_fcc_experiments}
%\centering
\resizebox{0.98\columnwidth}{!}{
\begin{tabular}{cccccc}
\hline
\thead{output\\ activation} & \thead{loss\\ function} & \thead{first\\ BN layer} & \thead{final\\ BN layer} & \thead{class-3\\ (minority)} & \thead{class-8\\ (majority)} \\\hline
sigmoid           & BCE           &        &        & 0.17                                   & 0.67                                   \\
softmax           & BCE           &        &        & 0.00                                      & 0.67                                   \\
softmax           & BCE           & \checkmark      &       & 0.60                                    & 0.78                                   \\
softmax           & CCE           &       &       & 0.67                                   & 0.80                                    \\
sigmoid           & BCE           &       & \checkmark      & 0.05                                   & 0.67                                   \\
softmax           & BCE           &        & \checkmark      & 0.85                                   & 0.88                                   \\
softmax           & BCE           & \checkmark      & \checkmark      & 0.83                                   & 0.87                                   \\
softmax           & CCE           &        & \checkmark      & \textbf{0.88}                          & \textbf{0.90}                           \\
softmax           & CCE           & \checkmark      & \checkmark      & 0.78                                   & 0.85\\                 
\hline
\end{tabular}
}
%\vspace{-4mm}
\end{table}

\section{Discussion}
\label{sec:discussion}
In a previous work done in ~\citep{kocaman2020improving}, the authors suggested that by applying BN to dense layers, the gap between activations is reduced (normalized) and then softmax is applied on normalized outputs, which are centered around the mean. Therefore, ending up with centered probabilities (around 0.5), but favoring the minority class by a small margin. They also argued that DNNs have the tendency of becoming ‘over-confident’ in their predictions during training, and this can reduce their ability to generalize further and thus perform as well on unseen data. Then they also concluded that a DNN with the final BN layer is more calibrated \citep{guo2017calibration}. We assert that 'being less confident' in terms of softmax outputs might be fundamentally wrong and a network shouldn't be discarded or embraced due to its capacity of producing confident or less confident results in the softmax layer as it may not even be interpreted as a 'confidence'.
%(the probability associated with the predicted class label reflects its ground truth correctness likelihood).

In this study, we empirically demonstrated that the final BN layer could still be eliminated in inference without compromising the attained performance gain. This finding supports the assertion that adding the BN layer makes the optimization landscape significantly smoother, which in turn renders the gradients' behavior more predictive and stable -- as suggested by \citep{santurkar2018does}. We argue that the learned parameters, which were affected by the addition of the final BN layer under imbalanced settings, are likely sufficiently robust to further generalization on the unseen samples, even without normalization prior to the softmax layer.

The observation of locating a sweet spot (10\%) for the imbalance ratio, at which we can utilize the final BN layer, might be explained by the fact that the backbone ResNet architecture is already strong enough to easily generalize on the PV dataset and the model does not need any other regularization once the number of samples from the minority class exceeds a certain threshold. As the threshold found in our experiments is highly related to the DL architecture and the utilized dataset, it is clear that it may not apply to other datasets, but can be found in a similar way.

As discussed before, the BN layer calculates mean and variance to normalize the previous outputs across the batch, whereas the accuracy of this statistical estimation increases as the batch size grows. However, its role seems to change under the imbalanced settings -- we found out that a batch size of 64 reaches the highest score, whereas by utilizing larger batches the score consistently drops.
As a possible explanation for this observation, we think that the larger the batches, the higher the number of majority samples in a batch and the lesser the chances that the minority samples are fairly represented, resulting in a deteriorated performance.

During our experiments, we expected to see similar behavior with sigmoid activation replacing softmax in the output layer, but, evidently, the final BN layer works best with softmax activations. Although softmax output may not serve as a good uncertainty measure for DNNs compared to sigmoid layer, it can still do well on detecting the under-represented samples when used with the final BN layer.

\section{Conclusions}
\label{sec:conclusion}
%This study illustrates that putting an additional BN layer just before the output layer has a considerable impact in terms of minimizing the training time and test error for minority classes in highly imbalanced datasets. 

%As evident from Table~\ref{tab:three_plants}, upon adding the final BN layer the F1 test score is increased from 0.2942 to 0.9562 for the unhealthy Apple minority class, from 0.7237 to 0.9575 for the unhealthy Pepper and from 0.5688 to 0.9786 for the unhealthy Tomato when WL is not used (all are averaged values over 10 runs).  When compared to a network with WL but without a final BN layer, the averaged improvements were 0.1615, 0.0636 and 0.1115 respectively. In short, we observed that the highest gain in test F1 score for both classes (majority vs.~minority) is achieved just by adding a final BN layer, resulting in a more than three-fold performance boost on some configurations.

In this study, we extended the previous efforts done by ~\citep{kocaman2020improving} to devise an effective approach to enable learning of minority classes, given the surprising evidence of applying the final BN layer. 
Given these recent findings, we formulated and tested additional hypotheses.% and report our observations throughout the paper.

We at first noticed that the performance gain after adding the final BN layer in highly imbalanced settings could still be achieved after removing this additional BN layer during inference; in turn enabling us to get a performance boost with no additional cost in production. Then we explored the dynamics of using the final BN layer as a function of the imbalance ratio within the training set, and found out that the impact of final BN on highly imbalanced settings is the most apparent when the ratio of minority class to the majority is less than 10\%; there is hardly any impact above that threshold. 

We also ran similar experiments with simpler architectures, namely a basic CNN and a single-layered FC network, when applied to the MNIST dataset under various imbalance settings. The simple CNN experiments exhibited a gain of $\sim$ 20\% boost per the minority class and $\sim$ 10\% per the majority class after adding the final BN layer. In the FC network experiments, we observed improvements by 10\% to 30\% only when a single majority class was defined; no improvements were evident for two majority classes, regardless of the usage of the final BN layer. While experimenting with different activation and cost functions, we found out that using the final BN layer with sigmoid activation had almost no impact on the task at hand.

We found the impact of final BN layer in simpler neural networks quite surprising. 
It is an important finding, which we plan to further investigate in the future, as it requires thorough analysis. We also plan to formulate our findings in a generalized way for any neural model, preferably with a combination of softmax activation or a proper loss function that could be used in imbalanced image classification problems.

%\section*{Acknowledgements}
%This research was supported by the Ministry of Science \& %Technology, Israel.

\bibliographystyle{plainnat}
\bibliography{References}

\begin{thebibliography}{21}
\providecommand{\natexlab}[1]{#1}
\providecommand{\url}[1]{\texttt{#1}}
\expandafter\ifx\csname urlstyle\endcsname\relax
  \providecommand{\doi}[1]{doi: #1}\else
  \providecommand{\doi}{doi: \begingroup \urlstyle{rm}\Url}\fi

\bibitem[Barz and Denzler(2020)]{barz2020deep}
Bjorn Barz and Joachim Denzler.
\newblock Deep learning on small datasets without pre-training using cosine
  loss.
\newblock In \emph{The IEEE Winter Conference on Applications of Computer
  Vision}, pages 1371--1380, 2020.

\bibitem[Beggel et~al.(2019)Beggel, Pfeiffer, and Bischl]{beggel2019robust}
Laura Beggel, Michael Pfeiffer, and Bernd Bischl.
\newblock Robust anomaly detection in images using adversarial autoencoders.
\newblock \emph{arXiv preprint arXiv:1901.06355}, 2019.

\bibitem[Bjorck et~al.(2018)Bjorck, Gomes, Selman, and
  Weinberger]{bjorck2018understanding}
Nils Bjorck, Carla~P Gomes, Bart Selman, and Kilian~Q Weinberger.
\newblock Understanding batch normalization.
\newblock In \emph{Advances in Neural Information Processing Systems}, pages
  7694--7705, 2018.

\bibitem[Chelombiev et~al.(2019)Chelombiev, Houghton, and
  O'Donnell]{chelombiev2019adaptive}
Ivan Chelombiev, Conor Houghton, and Cian O'Donnell.
\newblock Adaptive estimators show information compression in deep neural
  networks.
\newblock \emph{arXiv preprint arXiv:1902.09037}, 2019.

\bibitem[Cui et~al.(2019)Cui, Jia, Lin, Song, and Belongie]{cui2019class}
Yin Cui, Menglin Jia, Tsung-Yi Lin, Yang Song, and Serge Belongie.
\newblock Class-balanced loss based on effective number of samples.
\newblock In \emph{Proceedings of the IEEE Conference on Computer Vision and
  Pattern Recognition}, pages 9268--9277, 2019.

\bibitem[Goodfellow et~al.(2016)Goodfellow, Bengio, and
  Courville]{goodfellow2016deep}
Ian Goodfellow, Yoshua Bengio, and Aaron Courville.
\newblock \emph{Deep learning}.
\newblock MIT press, 2016.

\bibitem[Guo et~al.(2017)Guo, Pleiss, Sun, and Weinberger]{guo2017calibration}
Chuan Guo, Geoff Pleiss, Yu~Sun, and Kilian~Q Weinberger.
\newblock On calibration of modern neural networks.
\newblock In \emph{Proceedings of the 34th International Conference on Machine
  Learning-Volume 70}, pages 1321--1330. JMLR. org, 2017.

\bibitem[He et~al.(2016)He, Zhang, Ren, and Sun]{he2016deep}
Kaiming He, Xiangyu Zhang, Shaoqing Ren, and Jian Sun.
\newblock Deep residual learning for image recognition.
\newblock In \emph{Proceedings of the IEEE conference on computer vision and
  pattern recognition}, pages 770--778, 2016.

\bibitem[Hussain et~al.(2018)Hussain, Bird, and Faria]{hussain2018study}
Mahbub Hussain, Jordan~J Bird, and Diego~R Faria.
\newblock A study on {CNN} transfer learning for image classification.
\newblock In \emph{UK Workshop on Computational Intelligence}, pages 191--202.
  Springer, 2018.

\bibitem[Ioffe and Szegedy(2015)]{ioffe2015batch}
Sergey Ioffe and Christian Szegedy.
\newblock Batch normalization: Accelerating deep network training by reducing
  internal covariate shift.
\newblock \emph{arXiv preprint arXiv:1502.03167}, 2015.

\bibitem[Kocaman et~al.(2020)Kocaman, Shir, and B{\"a}ck]{kocaman2020improving}
Veysel Kocaman, Ofer~M Shir, and Thomas B{\"a}ck.
\newblock Improving model accuracy for imbalanced image classification tasks by
  adding a final batch normalization layer: An empirical study.
\newblock \emph{arXiv preprint arXiv:2011.06319, Accepted to International
  Conference on Pattern Recognition, ICPR 2020.}, 2020.

\bibitem[Krogh and Hertz(1992)]{krogh1992simple}
Anders Krogh and John~A Hertz.
\newblock A simple weight decay can improve generalization.
\newblock In \emph{Advances in neural information processing systems}, pages
  950--957, 1992.

\bibitem[Lake et~al.(2015)Lake, Salakhutdinov, and Tenenbaum]{lake2015human}
Brenden~M Lake, Ruslan Salakhutdinov, and Joshua~B Tenenbaum.
\newblock Human-level concept learning through probabilistic program induction.
\newblock \emph{Science}, 350\penalty0 (6266):\penalty0 1332--1338, 2015.

\bibitem[Mishkin and Matas(2015)]{mishkin2015all}
Dmytro Mishkin and Jiri Matas.
\newblock All you need is a good init.
\newblock \emph{arXiv preprint arXiv:1511.06422}, 2015.

\bibitem[Mohanty et~al.(2016)Mohanty, Hughes, and
  Salath{\'e}]{mohanty2016using}
Sharada~P Mohanty, David~P Hughes, and Marcel Salath{\'e}.
\newblock Using deep learning for image-based plant disease detection.
\newblock \emph{Frontiers in plant science}, 7:\penalty0 1419, 2016.

\bibitem[M{\"u}ller et~al.(2019)M{\"u}ller, Kornblith, and
  Hinton]{muller2019does}
Rafael M{\"u}ller, Simon Kornblith, and Geoffrey~E Hinton.
\newblock When does label smoothing help?
\newblock In \emph{Advances in Neural Information Processing Systems}, pages
  4696--4705, 2019.

\bibitem[Santurkar et~al.(2018)Santurkar, Tsipras, Ilyas, and
  Madry]{santurkar2018does}
Shibani Santurkar, Dimitris Tsipras, Andrew Ilyas, and Aleksander Madry.
\newblock How does batch normalization help optimization?
\newblock In \emph{Advances in Neural Information Processing Systems}, pages
  2483--2493, 2018.

\bibitem[Shorten and Khoshgoftaar(2019)]{shorten2019survey}
Connor Shorten and Taghi~M Khoshgoftaar.
\newblock A survey on image data augmentation for deep learning.
\newblock \emph{Journal of Big Data}, 6\penalty0 (1):\penalty0 60, 2019.

\bibitem[Simonyan and Zisserman(2014)]{simonyan2014very}
Karen Simonyan and Andrew Zisserman.
\newblock Very deep convolutional networks for large-scale image recognition.
\newblock \emph{arXiv preprint arXiv:1409.1556}, 2014.

\bibitem[Smith(2017)]{smith2017cyclical}
Leslie~N Smith.
\newblock Cyclical learning rates for training neural networks.
\newblock In \emph{2017 IEEE Winter Conference on Applications of Computer
  Vision (WACV)}, pages 464--472. IEEE, 2017.

\bibitem[Zhang et~al.(2017)Zhang, Cisse, Dauphin, and
  Lopez-Paz]{zhang2017mixup}
Hongyi Zhang, Moustapha Cisse, Yann~N Dauphin, and David Lopez-Paz.
\newblock mixup: Beyond empirical risk minimization.
\newblock \emph{arXiv preprint arXiv:1710.09412}, 2017.

\end{thebibliography}

\end{document}